\documentclass[11pt]{article}

\PassOptionsToPackage{table}{xcolor}

\usepackage{acl}

\usepackage{times}
\usepackage{latexsym}
\usepackage{amsmath}
\usepackage[most]{tcolorbox}
\usepackage{etoolbox}
\usepackage{lipsum}
\tcbuselibrary{listingsutf8}
\usepackage{xcolor}
\usepackage{colortbl}
\usepackage{hyperref}
\usepackage{ulem}

\usepackage{float}
\usepackage[T1]{fontenc}

\usepackage[utf8]{inputenc}

\usepackage{microtype}

\usepackage{inconsolata}

\usepackage{graphicx}
\usepackage{booktabs}
\usepackage{multirow}
\usepackage{array}
\usepackage{wrapfig}

\usepackage[linesnumbered,ruled,vlined]{algorithm2e}

\title{Improving Factuality in LLMs via Inference-Time \\Knowledge Graph Construction}

\author{Shanglin Wu \\
  Emory University\\
  \texttt{shanglin.wu@emory.edu} \\\And
  Lihui Liu \\
  Wayne State University\\
  \texttt{hw6926@wayne.edu} \\\And
  Jinho D. Choi \\
  Emory University\\
  \texttt{jinho.choi@emory.edu} \\\And
  Kai Shu \\
  Emory University\\
  \texttt{kai.shu@emory.edu} \\}

\tcbset{
  aclprompt/.style={
    colback=blue!5,
    colframe=blue!50,
    fonttitle=\bfseries,
    boxrule=0.6pt,
    left=4pt,
    right=4pt,
    top=10pt,
    bottom=10pt,
    breakable,
    enhanced,
    sharp corners,
    before skip=5pt,
    after skip=5pt,
    boxsep=0pt
  }
}

\begin{document}
\maketitle
\begin{abstract}
Large Language Models (LLMs) often struggle with producing factually consistent answers due to limitations in their parametric memory. Retrieval-Augmented Generation (RAG) paradigms mitigate this issue by incorporating external knowledge at inference time. However, such methods typically handle knowledge as unstructured text, which reduces retrieval accuracy, hinders compositional reasoning, and amplifies the influence of irrelevant information on the factual consistency of LLM outputs.
To overcome these limitations, we propose a novel framework that dynamically constructs and expands knowledge graphs (KGs) during inference, integrating both internal knowledge extracted from LLMs and external knowledge retrieved from external sources. Our method begins by extracting a seed KG from the question via prompting, followed by iterative expansion using the LLM's internal knowledge. The KG is then selectively refined through external retrieval, enhancing factual coverage and correcting inaccuracies.
We evaluate our approach on three diverse Factual QA benchmarks, demonstrating consistent gains in factual accuracy over baselines. Our findings reveal that inference-time KG construction is a promising direction for enhancing LLM factuality in a structured, interpretable, and scalable manner. The code for this paper is available at \url{https://github.com/ShanglinWu/autokg.git}.
\end{abstract}
\begin{figure}[!t]
  \centering
  \includegraphics[width=1.0\linewidth]{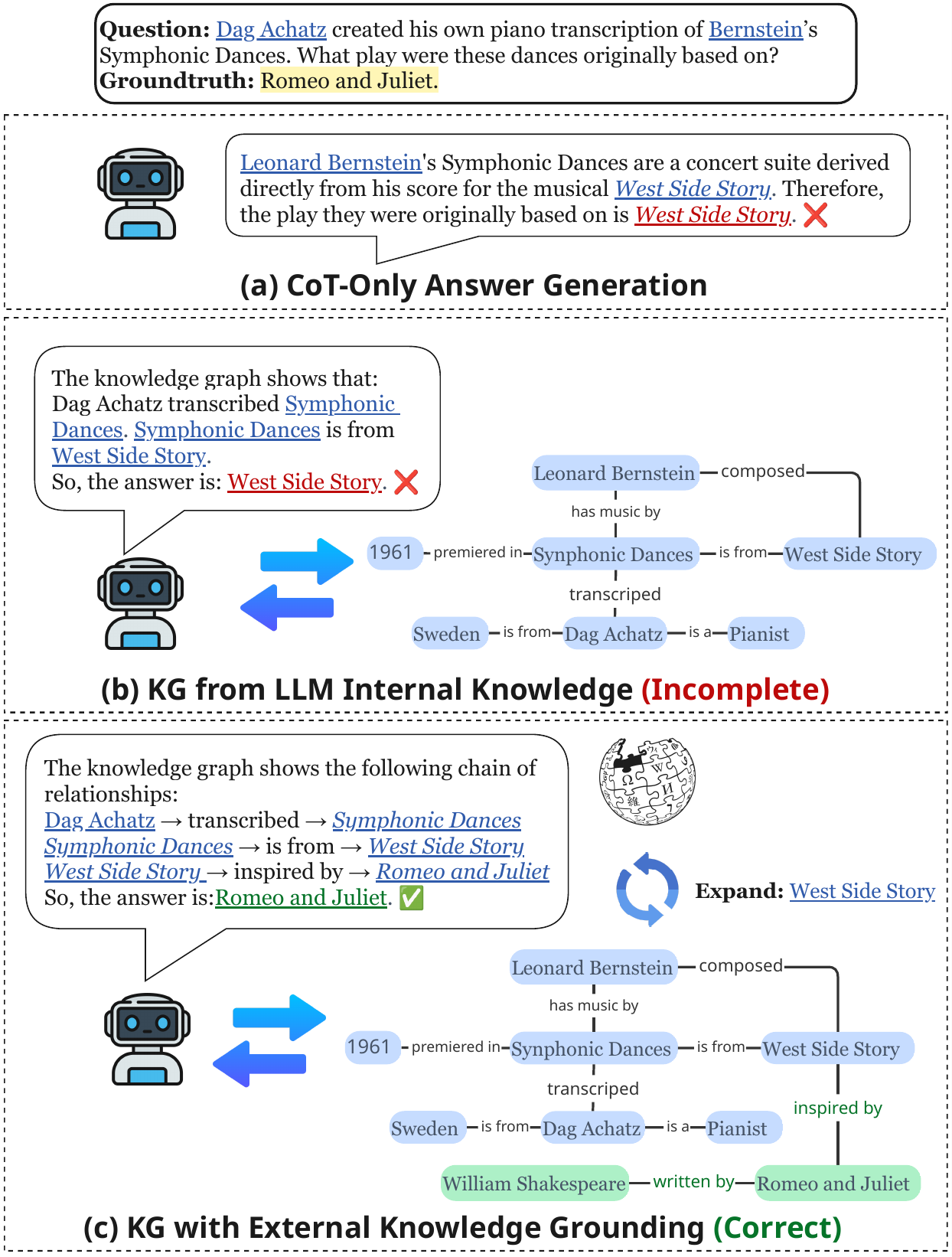}
  \caption{
    Comparison of three methods for answering factual questions: 
    (a) Chain-of-Thought prompting, 
    (b) Answering with a KG constructed from the LLM’s internal knowledge, 
    (c) Answering with a KG grounded in external knowledge. 
    Grounding with external knowledge corrects incorrect entities and expands KG with new edges, ultimately leading to the correct answer. 
  }
  \label{fig:kg_comparison}
\end{figure}

\section{Introduction}

Large Language Models (LLMs) have demonstrated remarkable capabilities in natural language understanding \cite{karanikolas2023large, kumar2024large}, content generation \cite{moore2023empowering, nejjar2023llms}, and reasoning \cite{saparov2022language,feng2023language,hao2024llm,an2023learning}. However, their propensity to generate content inconsistent with factual information presents substantial risks in high-stakes domains and degrades overall performance.\cite{wang2024factuality,wang2024openfactcheck}. This issue have been highlighted by evaluating LLMs on factual question answering tasks, where LLMs are expected to provide accurate answers to questions based on verifiable facts. Resent research indicates that even for relatively straightforward factual retrieval tasks, LLMs struggle with reliably grounding their responses \cite{huang2025survey}. 

To address aforementioned factuality concerns, researchers have increasingly turned to Retrieval-Augmented Generation (RAG) approaches that supplement LLMs with external evidence at inference time \cite{zhao2024retrieval,lee2025rearag}. While RAG has emerged as a promising solution to factuality challenges, most existing approaches treat retrieved knowledge as unstructured text. However, recent research shows that text-only retrieval suffers from lower retrieval accuracy, limited compositional reasoning, and heightened sensitivity to irrelevant evidence, which together undermine factual consistency\cite{han2024retrieval, wang2025retrieval}. 

To this end, recent works demonstrate the effectiveness of merging knowledge graphs (KGs) at inference time to enhance LLMs' factuality  \cite{pan2024unifying, yao2023beyond, sun2023think}. KGs represent knowledge as interconnected triplets, offering explicit structure that supports compositional reasoning, and mitigates hallucination. Instead of relying purely on unstructured parametric memory, reasoning over structured KG enables LLMs to access and integrate factual knowledge systematically. Furthermore, the graph-based representation of knowledge facilitates the identification of knowledge deficiencies \cite{zheng2023can}, demonstrating the potential to refine the KG by LLMs precisely during inference time.

Despite these advantages, most existing methods still depend on static, pre-constructed KGs as their main factual sources, which constrains adaptability. Such methods depend on curated domain knowledge yet remain difficult to maintain and scale in fast-changing contexts\cite{sen2023knowledge,baek2023knowledge,wu2023retrieve}. In contrast, LLMs trained on massive text corpora possess broad but implicit world knowledge\cite{gekhman2025inside}, allowing on-demand knowledge extraction and dynamic KG construction with minimal manual effort. This contrast highlights two complementary paradigms: one that relies on externally curated, static KGs, and another that builds dynamic, LLM-derived KGs at inference time. This naturally raises the question:
\textbf{Can we combine the strengths of both paradigms leveraging the customizability and reliability of external knowledge bases with the adaptability and efficiency of LLM-derived KGs?} Such a hybrid approach could enable real-time KG construction and expansion during inference-time, thereby improving LLMs’ factuality and reliability. 

To illustrate the core motivation and methodological differences, Figure~\ref{fig:kg_comparison} compares three approaches to factual question answering: (a) Chain-of-Thought (CoT) prompting; (b) answering with a KG constructed from the LLM’s internal knowledge; and (c) answering with a KG further refined through external retrieval. Compared to unstructured CoT reasoning, KGs make the reasoning process more explicit and interpretable. Moreover, incorporating external knowledge allows the model to revise missing or inaccurate triplets, thereby improving factual coverage and answer accuracy.

We propose a novel pipeline that constructs and expands KGs dynamically during inference-time, drawing from both internal LLM representations and external knowledge bases. By grounding intermediate reasoning in a graph-structured form, our approach enhances answer accuracy, interpretability (via explicit evidence chains), and robustness to missing or spurious context, while remaining plug-and-play across backbone models and knowledge sources. Extensive experiments on multiple Question Answering(QA) benchmarks demonstrate consistent gains. Our main contributions are:
\begin{itemize}
    \item \textbf{Inference-Time KG Construction:} We present a novel method to extract triplets directly from LLMs' internal knowledge, design an inference-time expansion strategy that iteratively extends KGs.

    \item \textbf{Hybrid Knowledge Integration:} We explore a unified pipeline that seamlessly combines LLMs' internal knowledge with external knowledge via KGs. This hybrid setup enhances factual grounding by effectively compensating for gaps in each source.

    \item \textbf{Empirical Validation on Factual QA Tasks:} We evaluate our method on multiple Factual Question Answering benchmarks, demonstrating consistent and significant improvements in answer accuracy and factual consistency across diverse datasets. The method remains robust under different backbone models, indicating strong generalization ability.
\end{itemize}

\section{Related Work}
\label{factualqa}
\noindent \textbf{Construct Knowledge Graphs by LLMs}
Recent advances in LLMs have led to growing interest in using them for knowledge graph construction. Traditional KG construction is a labor-intensive task requiring extensive domain expertise, whereas LLMs offer the potential to automate this process by leveraging their pre-trained knowledge and natural language understanding.  \citet{kommineni2024human} introduce a semi-automatic pipeline where an LLM generates competency questions to guide ontology construction and triplet extraction from academic texts. \citet{nie2024knowledge} employ chain-of-thought prompting to improve relation extraction consistency using predefined schemas, while enhancing graph quality through ontology alignment. \citet{xu2024generate} present Generate-on-Graph (GoG), a training-free method where an LLM dynamically augments incomplete KGs during question answering by injecting new triplets based on its world knowledge. For data-scarce settings,  \citet{li2024llm} propose FKGC, which synthesizes plausible triplets for rare relations and distills them into smaller models with consistency checks.

LLM-based KG construction has also been explored in various domain-specific and QA-driven contexts.  \citet{bai2025construction} construct a 2-million-node KG (KG-FM) from over 100,000 research papers using Qwen2-72B, enabling accurate QA through KG-augmented inference. Meanwhile, \citet{ding2024automated} propose pipelines for theme-specific and cybersecurity KGs, guided by domain-specific ontologies to enhance structure and accuracy. Complementing these text-based approaches, \citet{cohen2023crawling} propose LMCRAWL, a method for extracting knowledge graphs directly from the internal knowledge of LLMs. Starting from seed entities, the model is prompted to recursively generate related triplets, building a structured KG without relying on external corpora. Their results show that LLMs encode a substantial amount of factual knowledge that can be surfaced with high precision. Compared with these methods, our approach automates the entire KG construction process, starting directly from plain-text input while maintaining high factual accuracy. This design makes the system both efficient and easy to use.

\noindent \textbf{Factual Question Answering}
Factual Question Answering(Factual QA) has been widely explored across different data sources and reasoning formats. Despite recent breakthroughs in LLMs, state-of-the-art models still face notable limitations in producing consistent and verifiable factual responses across QA benchmarks. These shortcomings have renewed interest in structured approaches such as KG-based QA to improve factual accuracy and reasoning transparency. Factual QA benchmarks are generally classified into three categories according to knowledge source: Knowledge Base QA (KBQA) focuses on answering questions from curated knowledge bases such as Freebase and Wikidata. Datasets like SimpleQuestions \cite{bordes2015large} and WebQuestionsSP \cite{yih2016value} provide natural language questions mapped to single-hop and multi-hop facts, while ComplexWebQuestions \cite{talmor2018web} and LC-QuAD 2.0 \cite{dubey2019lc} introduce complex logical, comparative, and numerical queries.  Document-based QA requires answering questions using evidence from unstructured text. HotpotQA \cite{yang2018hotpotqa} is a widely used multi-hop benchmark where models must synthesize information across multiple paragraphs, while ConditionalQA \cite{sun2022conditionalqa} includes questions whose answers depend on specific conditions mentioned in context. DROP \cite{dua2019drop} poses challenges involving numerical, comparative, and logical reasoning.  Expert-curated QA datasets, such as SimpleQA \cite{wei2024simpleqa}, which consists of simple fact-seeking questions with verifiable answers. Because the questions were adversarially collected against GPT-4, many are specifically tailored to exploit models’ blind spots, making it difficult even for SOTA LLMs to reliably answer them.

\noindent \textbf{Knowledge Graph–Augmented Factual Question Answering}
Moving from KG construction, recent works evaluate how KGs enhance LLM factuality in Factual QA tasks. \citet{pan2024unifying} propose a roadmap for unifying LLMs and KGs, arguing that structured, verified facts provided by KGs help mitigate hallucinations and ensure factual consistency. They demonstrate that retrieving from KGs supplements the parametric memory of LLMs and supports more accurate factual retrieval. \citet{yao2023beyond} and \citet{sun2023think} emphasize that the explicit graph structure enables compositional, multi-hop reasoning by linking related facts through traversable paths. Their studies show that reasoning over KGs naturally facilitates complex inference steps that are difficult for LLMs relying solely on free-text generation. Empirically, \citet{ma2025debate} introduce Debate-on-Graph (DoG), a KG-augmented reasoning framework, achieving a 23.7\% improvement in accuracy over prior methods on WebQuestions. Their results highlight that incorporating KG-based reasoning loops leads to more systematic, interpretable, and fact-grounded QA performance. Building on these advances, our approach unifies fully automatic, factually accurate KG construction with structure-aware external retrieval, achieving strong generalization across input formats and knowledge sources and surpassing prior methods in performance.

\section{Method}
\label{sec:method}
\begin{figure*}[!t]
  \centering
  \includegraphics[scale=0.37]{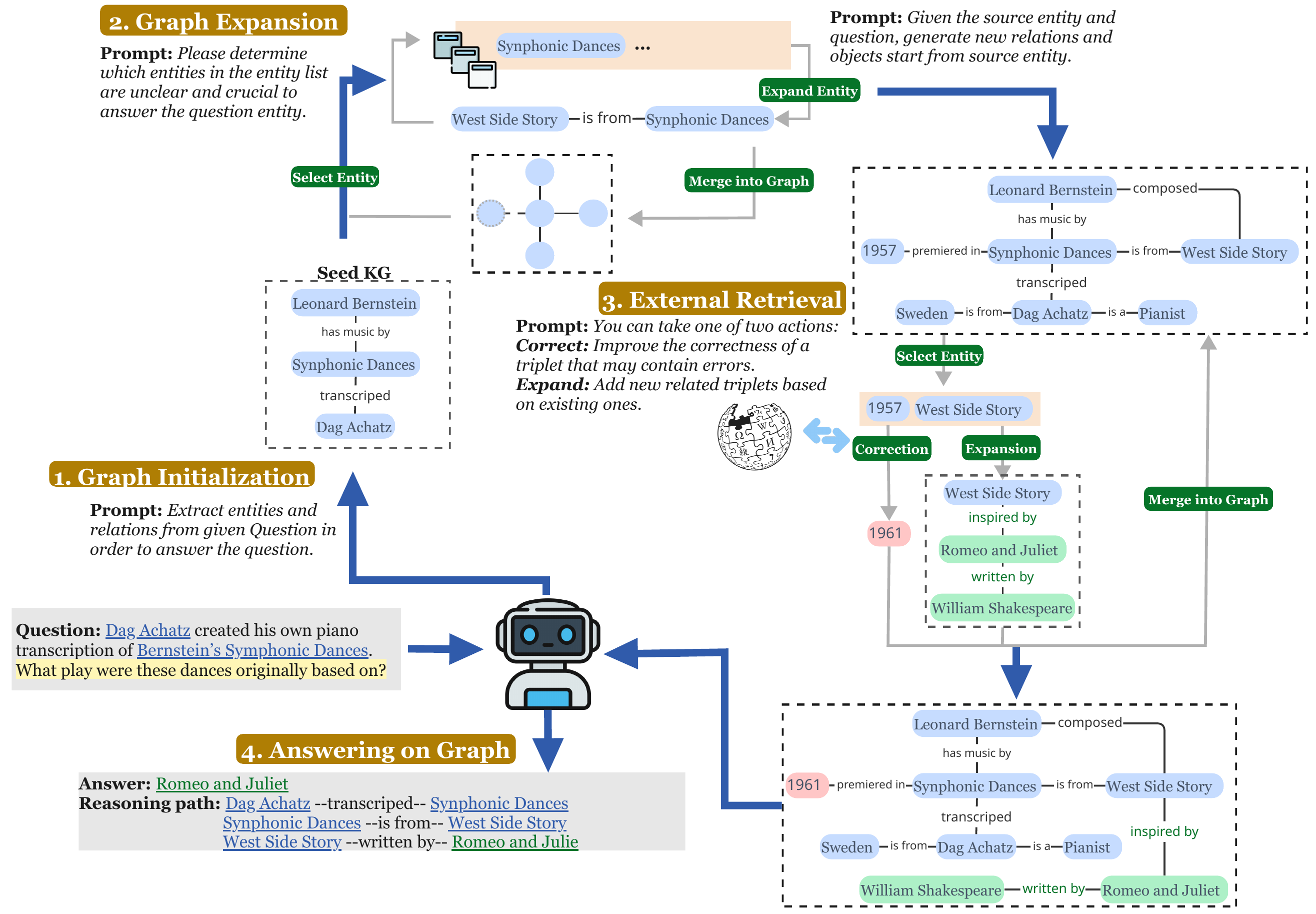}
  \caption{Overview of our pipeline. \textbf{(A) Graph initialization}, in which the input question is parsed by the LLM to extract initial triplets. \textbf{(B) Graph expansion} iteratively explores breadth‑first relations from seed entities to build a larger KG. \textbf{(C) External retrieval}, search is performed (e.g., using BM25 on the content returned from wikipedia and google search) to correct or extend selected triplets, which are merged into the graph. \textbf{(D) Answering on Graph}, a refined KG supports factual answer generation, yielding a grounded response.}
  \label{fig:pipeline}
\end{figure*}

In this section we introduce our method, a four-stage pipeline that enables LLMs to automatically construct, refine, and utilize knowledge graphs for Factual Question Answering. Formally, a KG is represented as $G = (s, r, o) \subseteq E \times R \times E$, where $E$ denotes the set of entities and $R$ represents the set of relations. Each triplet $(s, r, o)$ consists of a subject-relation-object structure, where $s, o \in E$ and $r \in R$. Additionally, we use $Q$ to represent a given question, and $LM$ denote a language model. 

As illustrated in Figure \ref{fig:pipeline}, the process can be formally described as follows: \textbf{Graph Initialization:} The process begins by initializing a seed KG $G_0$ from the given question $Q \rightarrow LM \rightarrow G_0$, which includes the entities and relations mentioned within the question.
\textbf{Graph Expansion:} In this step, the seed KG $G_0$ is expanded in a breadth-first order. For each iteration, the LLM generates a new relation $r$ and object $o$ from a selected source entity $s$. This expansion continues until the KG reaches a depth of $D$, resulting in the expanded KG $G_1$.
\textbf{External Retrieval:} When leveraging the $G_1$ to answer questions, LLM may encounter incorrect or incomplete triplets, either containing factually incorrect entities (incorrect) or lacking sufficient detail and thus requiring further expansion to reach the correct answer (incomplete). To address this, the LLM corrects or expands $G_1$, such as replacing an erroneous triplet $(s, r, o)$ with a corrected triplet $(s, r', o')$ or add new triplets $\{(s, r^*_1,o^*_1), (o^*_
1, r^*_2,o^*_2)...\}$ to $G_1$. During this stage, the LLM has access to external knowledge sources. The revised KG, incorporating external information, is denoted as $G^*$.
\textbf{Answering on Graph:} The final step involves the LLM generating an answer to the question $Q$ by querying the relations and entities within $G^*$. 




\subsection{Graph Initialization}
Given a question $Q$, the LLM is first prompted to extract relevant entities and relations, returns a set of triplets that capture the core focus of $Q$. These triplets are utilized to initialize the seed KG $G_0$, which forms the basis for the subsequent expansion process. For example, in Figure \ref{fig:pipeline}, when given the question \textit{"Dag Achatz created his own piano transcription of Bernstein’s Symphonic Dances — what play were these dances originally based on?”}, the LLM extracts entities \textit{\{Dag Achatz, Symphonic Dances, Leonard Bernstein\}}, forming $G_0$
.
\subsection{Graph Expansion}
This stage forms the core mechanism for constructing a KG from the LLM's internal knowledge. We design an iterative process in which the LLM selects a subset of entities deemed valuable for further exploration. These selected entities, denoted as $E_s$, are placed into a buffer $B$. The LLM then proceeds to explore each entity in $B$ using a breadth-first strategy. During each iteration, the LLM generates one new triplet related to the current entity. This process continues until the buffer is empty or a predefined depth $D$ limit is reached. The resulting expanded KG is denoted as $G_1$. Building on the previous illustrative example, the LLM expands from \textit{Symphonic Dances} to generate new triplet \textit{(Symphonic Dances, is from, West Side Story)}, gradually linking intermediate entities that connect the initial concepts to the underlying play.

\subsection{External Retrieval}
Although the LLM can attempt to answer the original question $Q$ using only the information encoded in the knowledge graph $G_1$, its internal knowledge base may be insufficient to generate a fully accurate answer. To address this limitation, we introduce an external retrieval mechanism that incorporates knowledge from Wikipedia and Google Search. When the expansion produces incomplete or uncertain facts (e.g., whether \textit{West Side Story} was inspired by another work), the external retrieval step searches trusted sources such as Wikipedia, correcting the KG by adding \textit{(West Side Story, inspired by, Romeo and Juliet)}. The LLM is allowed to iteratively improve the KG up to a limited number of times. In each iteration, the LLM selects an entity or triplet from the KG and chooses one of two possible actions: \textbf{Correction} or \textbf{Expansion}. \textbf{Correction} refers to improving the accuracy of a triplet that may contain errors, while \textbf{Expansion} involves adding new related triplets based on the existing ones. Specifically, we construct a search query from the selected triplet $(s, r, o)$ and retrieve the top $k$ most relevant search results using the BM25 algorithm \cite{robertson2009probabilistic}. The search result is then provided to the LLM, which returns the corrected or expanded triplet(s), which are merged into the KG. By leveraging reliable and diverse external information, LLM refines and updates the KG, resulting in an improved KG $G^*$. 

\subsection{Answering on the Graph}

Finally, the LLM is prompted to generate an answer to the question $Q$ based on $G^*$. This structured representation ensures that the LLM’s response is grounded in verifiable facts while also exposing the reasoning paths on the KG. As shown in Figure \ref{fig:pipeline}, the LLM reasons over $G^*$ to trace a factual path \textit{Dag Achatz → Symphonic Dances → West Side Story → Romeo and Juliet} and outputs \textit{Romeo and Juliet} as the final answer. We restrict model to triplets present in $G^*$ to reduce off-graph hallucinations and improve faithfulness. All prompts used during this process are provided in Appendix \ref{app:prompt}.

\begin{table*}[htbp!]
\centering
\small
\renewcommand{\arraystretch}{1.2}
\setlength{\tabcolsep}{6pt}
\begin{tabular}{@{}l ccc ccc cc@{}}
\toprule
\textbf{Method} 
& \multicolumn{3}{c}{\textbf{CWQ}} 
& \multicolumn{3}{c}{\textbf{HotpotQA}} 
& \multicolumn{2}{c}{\textbf{SimpleQA}} \\
\cmidrule(lr){2-4} \cmidrule(lr){5-7} \cmidrule(lr){8-9}
& Acc. & EM & Recall & Acc. & EM & Recall & Acc. & Recall \\
\midrule

\multicolumn{9}{c}{\textbf{Internal Knowledge}} \\
\midrule
\quad CoT \cite{wei2022chain}   & 64.4 & 46.7 & 46.5 & 45.3 & 36.1 & 31.8 & 11.7 & 7.9 \\
\quad GoT \cite{besta2024graph}   & 67.3 & 48.0 & 53.6 & 47.4 & 37.4 & 37.7 & 9.5 & 6.7 \\
\quad Ours  & \textcolor{green!50!black}{\underline{68.1}} & \textcolor{green!50!black}{\underline{52.5}} & \textcolor{green!50!black}{\underline{59.1}} & \textcolor{green!50!black}{\underline{48.3}} & \textcolor{green!50!black}{\underline{41.1}} & \textcolor{green!50!black}{\underline{43.6}} & \textcolor{green!50!black}{\underline{14.1}} & \textcolor{green!50!black}{\underline{13.0}} \\

\midrule
\multicolumn{9}{c}{\textbf{External Knowledge}} \\
\midrule
\quad TextRAG & 61.5 & 45.3 & 48.6 & 48.3 & 37.1 & 36.2 & 16.6 & 13.1 \\
\quad ToG \cite{sun2023think}     & 66.0 & 48.5 & 56.7 & 48.6 & 37.0 & 39.4 & 10.7 & 9.2 \\
\quad GoG \cite{xu2024generate}  & 67.4 & 53.5 & \textcolor{green!50!black}{\underline{65.7}} & 46.9 & 36.4 & 36.3 & 10.5 & 8.7 \\
\quad Ours    & \textcolor{green!50!black}{\underline{68.7}} & \textcolor{green!50!black}{\underline{55.2}} & 61.5 & \textcolor{green!50!black}{\underline{54.8}} & \textcolor{green!50!black}{\underline{43.0}} & \textcolor{green!50!black}{\underline{48.6}} & \textcolor{green!50!black}{\underline{37.1}} & \textcolor{green!50!black}{\underline{37.5}} \\

\bottomrule
\end{tabular}
\caption{Performance comparison with baselines, grouped by internal vs. external knowledge methods. The backbone model is Llama-4-Scout.}
\label{tab:ours_vs_baselines}
\end{table*}

\section{Experiments}

\subsection{Datasets}
The test questions should be simple and concise, with answers consisting of a single entity. This ensures that LLM can accurately interpret the question, allowing the evaluation to focus on factual grounding. Considering the diversity of knowledge sources and the broad coverage of topics and entities, we select three representative datasets for question answering across different categories: \textbf{Complex WebQuestions (CWQ)} \cite{talmor2018web}, \textbf{HotpotQA} \cite{yang2018hotpotqa}, and \textbf{SimpleQA} \cite{wei2024simpleqa}, encompassing categories such as Knowledge Base Question Answering (KBQA), document-based QA, and expert-curated QA.

\subsection{Baselines}
The baselines in our comparison can be grouped into two categories:  
(a) \textbf{Internal Knowledge:} methods that rely solely on the LLMs' internal knowledge. We consider Chain-of-Thought \cite{wei2022chain}prompting and Graph-of-Thoughts \cite{besta2024graph} as representatives baseline methods of this group.  
(b) \textbf{External Knowledge:} Methods that incorporate retrieval from external knowledge sources. This group includes text-based RAG, which leverages Wikipedia as the knowledge source and integrates it with LLM inference, as well as Think-on-Graph ~\cite{sun2023think} and Generate-on-Graph ~\cite{xu2024generate}.

\subsection{Experiment Settings}
In our experiments, we evaluate the performance of various models on CWQ-test  (3,519 questions), HotpotQA-dev (7,405 questions), and SimpleQA (4,326 questions). For HotpotQA-dev, we only provide questions to LLM without the support contexts. The evaluation metrics include Exact Match (EM) and LLM-assisted judge accuracy (Acc) on HotpotQA and CWQ, for SimpleQA we utilize simple-evals framework. For the LLM-assisted judge, we utilize GPT-4o-mini. For each dataset and method, we also report recall, which denotes the proportion of ground truth facts appearing in the reasoning process or retrieved context. The models under evaluation are GPT-4o, Deepseek-v3, Gemini-2.5-flash, Qwen2.5-32B-instruct, and Llama-4-scout. For external retrieval, we utilize Serper API for google search and Wikipedia python package. The maximum token length for each generation is set to 256. The temperature parameter is set to 0.7. For all experiments, we conduct multiple runs and report the averaged performance. The results of these evaluations are presented in Table \ref{tab:ours_vs_baselines} and Table \ref{tab:model_comparison}. Detailed prompts and experimental settings are provided in Appendix \ref{app:prompt} and Appendix \ref{app:hyper}.

\begin{table}[t!]
\centering
\small
\renewcommand{\arraystretch}{1.2}
\setlength{\tabcolsep}{5pt}
\begin{tabular}{@{}l cc cc c@{}}
\toprule
\textbf{Model} 
& \multicolumn{2}{c}{\textbf{CWQ}} 
& \multicolumn{2}{c}{\textbf{HotpotQA}} 
& \multicolumn{1}{c}{\textbf{SimpleQA}} \\
\cmidrule(lr){2-3} \cmidrule(lr){4-5} \cmidrule(lr){6-6}
& Acc. & EM & Acc. & EM & Acc. \\
\midrule

\multicolumn{6}{c}{\textbf{Internal Knowledge}} \\
\midrule
GPT\mbox{-}4o             & \textcolor{green!50!black}{\underline{75.3}} & \textcolor{green!50!black}{\underline{55.2}} & \textcolor{green!50!black}{\underline{66.2}} & \textcolor{green!50!black}{\underline{43.8}} & 24.9 \\
Gemini\mbox{-}2.5\mbox{-}flash & 63.4 & 48.4 & 63.4 & 37.6 & \textcolor{green!50!black}{\underline{25.3}} \\
Qwen\mbox{-}2.5\mbox{-}32B     & 63.9 & 43.9 & 46.5 & 33.7 & 10.6  \\
Llama\mbox{-}4\mbox{-}scout     & 68.1 & 52.5 & 48.3 & 41.1 & 14.1  \\
Deepseek\mbox{-}V3        & 73.9 & 54.6 & 63.8 & 43.1 & 24.6 \\
\midrule

\multicolumn{6}{c}{\textbf{External Knowledge}} \\
\midrule
GPT\mbox{-}4o             & \textcolor{green!50!black}{\underline{75.4}} & 56.2 & 67.6 & \textcolor{green!50!black}{\underline{45.0}} & \textcolor{green!50!black}{\underline{43.3}} \\
Gemini\mbox{-}2.5\mbox{-}flash & 70.8 & 54.4 & \textcolor{green!50!black}{\underline{68.2}} & 43.2 & 23.5 \\
Qwen\mbox{-}2.5\mbox{-}32B     & 65.4 & 45.7 & 47.8 & 35.8 & 21.3 \\
Llama\mbox{-}4\mbox{-}scout     & 68.7 & 55.2 & 54.8 & 42.1 & 37.1 \\
Deepseek\mbox{-}V3        & 74.8 & \textcolor{green!50!black}{\underline{56.3}} & 64.8 & 45.0 & 35.4 \\
\bottomrule
\end{tabular}
\caption{Performance comparison using different backbone models across three QA datasets. }
\label{tab:model_comparison}
\end{table}

\subsection{Main Results}
\subsubsection{Comparison to Other Methods}
Table~\ref{tab:ours_vs_baselines} presents the performance comparison between our methods and existing baselines. Our approach achieves state-of-the-art results under both the internal and external knowledge settings. Under the internal knowledge setting, our method attains the highest performance on CWQ and HotpotQA, with a notable improvement in recall. This indicates that the observed performance gain stems from our method’s ability to more effectively elicit relevant information from the LLM’s internal knowledge base. However, all three methods perform poorly on SimpleQA, revealing the inherent limitation of relying solely on an LLM’s internal knowledge. Under the external knowledge setting, our method substantially improves performance, particularly on SimpleQA, with an accuracy gain of up to 23.0\%. The superiority of our method can be attributed to two key factors: (a) Compared to text-based RAG, our approach retrieves information more precisely by identifying incorrect or missing knowledge pieces through triples on the KG, thereby enhancing retrieval efficiency and reducing the impact of irrelevant context. (b) Compared to other graph-based RAG methods (e.g., ToG and GoG), our approach offers greater flexibility in selecting external knowledge sources. While methods that rely on the same knowledge base as the dataset (e.g., GoG utilizing Freebase for CWQ) perform well in such cases, they struggle with datasets like SimpleQA, where the knowledge source is carefully human-curated. In contrast, our method’s ability to robustly integrate multiple heterogeneous knowledge sources proves crucial for complex real-world settings.

\subsubsection{Performances with Different Backbone Models}
We further evaluate how different backbone models influence overall performance. As shown in Table~\ref{tab:model_comparison}, performance generally improves as the model size increases, which aligns with our expectation. However, the performance gap between models becomes smaller under the \textit{external knowledge} setting, suggesting that the presence of external knowledge sources mitigates the limitations of the LLM’s internal knowledge base. Moreover, as model scale increases, the relative improvement brought by external retrieval tends to diminish, whereas for more challenging datasets, the benefit from external retrieval becomes more pronounced. Notably, on the SimpleQA dataset under the external knowledge setting, Llama-4-Scout (109B, 37.1\%) outperforms DeepSeek-V3 (671B, 35.4\%), and Qwen-2.5-32B (21.3\%) achieves comparable performance to Gemini-2.5-Flash (23.5\%). This finding highlights the potential of our method to enable smaller, more cost-efficient models to achieve performance on par with substantially larger ones.

\begin{figure*}[t]
  \centering
  \includegraphics[scale=0.4]{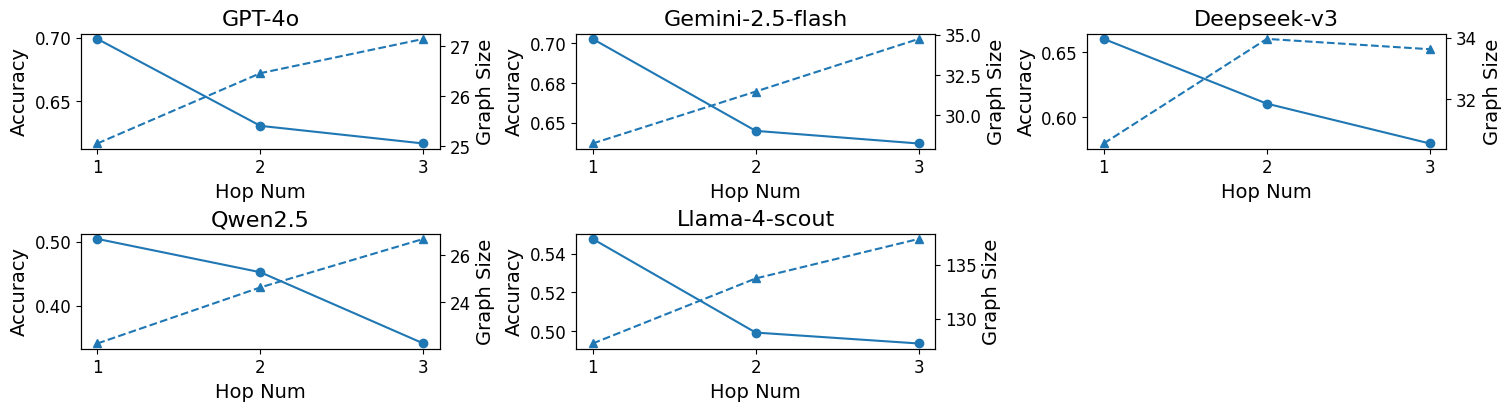}
  \caption{Accuracy and graph size for five models across different hop counts. Solid lines represent accuracy, while dashed lines indicate graph size.}
  \label{fig:hops}
\end{figure*}

To better understand the contribution of external retrieval in improving knowledge coverage of KGs, we compute the recall of the constructed graphs—i.e., the proportion of groundtruth appear in the graphs. The results are shown in Table~\ref{tab:recall_internal_external}. Across all datasets and models, the recall consistently improves when external sources are integrated. For instance, on the SimpleQA dataset, Llama-4-scout shows a substantial recall increase from 13.0\% to 37.5\%, a gain of +24.5 points. Similarly, Qwen2.5-32B improves by +27.0 points on SimpleQA. These gains are particularly prominent on SimpleQA, a dataset with sparse but precise fact requirements, which highlights the brittleness of relying solely on internal LLM knowledge. Even for stronger models like GPT-4o and Deepseek-V3, external retrieval leads to consistent improvements across all datasets. This finding underscores the generalization capability of our method and suggests that its performance gains are largely attributable to the extensive knowledge coverage provided by the KGs, while the plug-and-play design makes our approach easy to deploy in practice.

\begin{table}[t!]
\centering
\small
\renewcommand{\arraystretch}{1.1}
\setlength{\tabcolsep}{6pt}
\begin{tabular}{lccc}
\toprule
\textbf{Model} & \textbf{CWQ} & \textbf{HotpotQA} & \textbf{SimpleQA} \\
\midrule
\multicolumn{4}{l}{\textit{GPT-4o}} \\
\quad Internal KG  & 64.6 & 51.8 & 25.8 \\
\quad + External   & \cellcolor{blue!10}66.0 & \cellcolor{blue!10}53.9 & \cellcolor{blue!10}48.1 \\
\quad $\Delta$ & +1.4 & +2.1 & +22.3 \\
\midrule
\multicolumn{4}{l}{\textit{Gemini-2.5-flash}} \\
\quad Internal KG  & 63.5 & 52.0 & 22.6 \\
\quad + External   & \cellcolor{blue!10}67.8 & \cellcolor{blue!10}58.5 & \cellcolor{blue!10}49.4 \\
\quad $\Delta$ & +4.3 & +6.5 & +26.8 \\
\midrule
\multicolumn{4}{l}{\textit{Qwen2.5-32B}} \\
\quad Internal KG  & 52.2 & 35.4 & 10.5 \\
\quad + External   & \cellcolor{blue!10}53.2& \cellcolor{blue!10}41.7 & \cellcolor{blue!10}37.5 \\
\quad $\Delta$ & +1.0 & +6.3 & +27.0\\
\midrule
\multicolumn{4}{l}{\textit{LLama-4-scout}} \\
\quad Internal KG  & 59.1& 43.6 & 13.0 \\
\quad + External   & \cellcolor{blue!10}61.5& \cellcolor{blue!10}48.6 & \cellcolor{blue!10}37.5 \\
\quad $\Delta$ & +2.4 & +5.0 & +24.5 \\
\midrule
\multicolumn{4}{l}{\textit{Deepseek-V3}} \\
\quad Internal KG  & 69.0& 52.1 & 24.3 \\
\quad + External   & \cellcolor{blue!10}71.0 & \cellcolor{blue!10}54.2 & \cellcolor{blue!10}47.0 \\
\quad $\Delta$ & +2.0 & +2.1 & +22.7 \\
\bottomrule
\end{tabular}
\caption{Recall (\%) of KGs constructed from internal LLM knowledge versus those enhanced with external retrieval. The bottom row of each model shows the absolute gain from external retrieval.}
\label{tab:recall_internal_external}
\end{table}

\subsection{Ablation Study}

\textbf{Effect of number of hops}
HotpotQA is a multi-hop Factual QA dataset in which each question is annotated with the number of reasoning hops required to arrive at the correct answer. To further investigate how this reasoning complexity impacts performance, we analyze our method with external knowledge setting by examining both LLM-judged answer accuracy and the size of the KGs, measured by the number of triplets per graph. As shown in Figure~\ref{fig:hops}, we observe a clear trend: as the number of hops increases, the average graph size grows while the accuracy of the model's answers declines.

This behavior suggests that more complex questions, which require multi-hop reasoning across multiple facts, lead to generate larger initial KGs in an attempt to capture all potentially relevant entities and relations. These larger initial KGs, when further expanded and contribute to the overall increase in graph size. However, the expanded KG introduces more candidate paths and entities, which makes it increasingly difficult for the model to identify the correct answer node with high precision. The decline in accuracy reflects this challenge, revealing a trade-off between broader coverage through larger KGs and the precision required for accurate factual retrieval.

\begin{figure}[b!]
  \centering
  \includegraphics[scale=0.3]{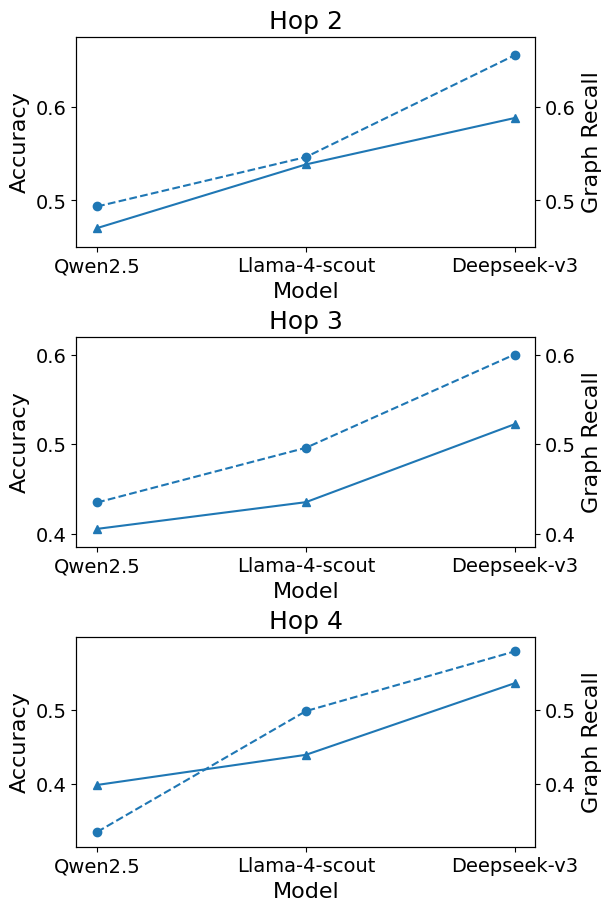}
  \caption{Accuracy and recall across different hop counts. Solid lines represent accuracy, while dashed lines indicate recall.}
  \label{fig:model_size}
\end{figure}

\textbf{Effect of model scale}
We further investigate the relationship between model size, accuracy, and recall, as shown in Figure~\ref{fig:model_size}. We evaluate three model scales—Qwen2.5 (32B), Llama-4-Scout (109B), and DeepSeek-V3 (671B). Overall, performance increases with model size: DeepSeek-V3 attains the highest accuracy and recall across all hop levels, highlighting the benefits of a larger internal knowledge base. Notably, it also exhibits the largest gap between accuracy and recall, suggesting that Deepseek-v3 relies more heavily on internal knowledge rather than the information explicitly contained in the graph. In contrast, Qwen2.5 performs the weakest overall. As the hop count increases, its accuracy even exceeds recall, indicating that it tends to rely on memorized knowledge rather than effectively leveraging graph structures. Llama-4-scout shows greater robustness to hop increase, indicating stronger capability in retrieving useful information from the KG in multi-hop scenarios.

\section{Conclusion}
In this paper, we present a novel pipeline for inference-time KG construction that integrates LLMs' internal knowledge and external knowledge. This approach improves factual accuracy and robustness in QA, with experiments showing its effectiveness in bridging knowledge gaps and remarkable generalization performance.

\section{Limitations}
Several limitations remain for our work: (a) When constructing knowledge graphs from the LLM’s internal knowledge, there is a risk of generating hallucinated content. Such noise can propagate through the graph and lead to incorrect answers. (b) A gap persists between graph recall and exact match performance, indicating that current graph retrieval mechanisms are still suboptimal. Improving this may further enhance overall performance.

\section{Ethical Consideration}
Potential benefits of our work include improved trustworthiness in applications like education, healthcare, and legal domains. However, misapplication might lead to overreliance on automated answers or misuse in misinformation contexts.  
We mitigate these risks by grounding responses in verifiable external sources and making reasoning chains interpretable. We also ensure that any dataset used complies with licensing and privacy regulations.  
All datasets used in our study are publicly released for academic research. HotpotQA is distributed under Apache License 2.0, Complex WebQuestions (CWQ) is released under the Creative Commons Attribution-ShareAlike (CC-BY-SA) license, and SimpleQA is distributed under the MIT License. We do not collect or redistribute any data from private or user-generated sources. Any content retrieved externally is used only at inference time for model augmentation and is not retained, repackaged, or released. Such use aligns with fair use principles for academic research.

\bibliography{custom}

\appendix

\section{Hyperparameters}
\label{app:hyper}

In our experiments, we adopt the following hyperparameters:

\begin{itemize}
    \item \textbf{Depth $D$}: The number of iterations for expanding the knowledge graph. In each iteration, the pipeline performs a breadth-first expansion on all seed entities in the queue, which are filtered from the current graph state. This reflects the maximum number of hops from entities extracted from the question to the outer boundary of the graph. In our experiments $D=3$.
    
    \item \textbf{Retrieval Steps $S$}: The number of iterations for external retrieval. In each step, the LLM selects one triplet from the graph and either modifies or expands it based on the retrieved external content. In our experiments $S=5$
    
    \item \textbf{BM25 Top-$k$ Paragraphs}: The number of top-ranked paragraphs returned by the BM25 algorithm for a given search query. These paragraphs are passed to the LLM as external context for enhancement or expansion. In our experiments $k=3$
\end{itemize}

\section{Prompts}
\label{app:prompt}
\subsection{LLM judgment}
\begin{tcolorbox}[title=LLM judgment, aclprompt]
You are an AI assistant for grading a factual question-answering task. You will receive a question, a model's prediction, and one or more correct answers. Your goal is to judge whether the prediction is factually aligned with any of the ground truths.

Do not require an exact wording match. Accept synonyms, abbreviations, paraphrases, or contextually correct answers. Even if the prediction is only part of the full ground truth, or includes extra information, consider it correct as long as the key factual content is accurate.

If multiple ground truths are given, matching any of them is enough. Use common sense and background knowledge. Only mark the prediction wrong if it clearly contradicts the facts.

Question:

[question]

Prediction:

[prediction]

Ground Truth(s):

[ground\_truth]

Explain your reasoning, and end your response on a new line with only "Yes" or "No" (without quotes)

\end{tcolorbox}

\subsection{Graph Construction}

\begin{tcolorbox}[title=Extract Question Entities, aclprompt]

Extract useful entities and relations from the given question to help answer it.

        Question: [question]

        Step 1 - Return new useful entities in this format:
        1. entity1
        2. entity2
        ...

        Step 2 - Then, extract any relationships **between those entities** based on your knowledge of the question.

        Return relations in this format:
        1. [entity] -> [relation] -> [entity]
        2. ...

        Requirements:
        
        - Only include relations among the extracted entities.
        
        - **Return only the lists**, no explanation or reasoning.

\end{tcolorbox}

\begin{tcolorbox}[title=Filter Entities, aclprompt]
Given a question, and an entity list. Please determine which entities in the entity list are unclear and crucial to answer the question.
        
        [Question]: [question]
        
        [Entity list]: [entities]
        
        You should return the entities strictly in the format of:
        
        [Trajectory]: your step-by-step thinking process.
        
        [Entity]
        1. entity1
        2. entity2
        ...

        Examples:
        
        Question: Which Emmett's Mark actor also played in the HBO series \"The Wire\"
        Entity list: ["Emmett's Mark", 'actor', 'The Wire', 'HBO']
        Knowledge Graph: Emmett's Mark --'has actor'--> actor\\nactor 'played in'--> HBO series "The Wire"
        Entity: 1. actor
        
        Question: Who directed the upcoming British action comedy film which has Johnny English as the first part? 
        Entity list: ["Johnny English", "Upcoming British action comedy film", "Director"]
        Knowledge Graph: Johnny English --is first part of'--> Upcoming British action comedy film\\nDirector --directs--> Upcoming British action comedy film
        Entity: 1. Director

        Question: In what year was the park, which hosts The Hot Ice Show, founded ?
        Entity list: ["park", "year", "Hot Ice Show"]
        Knowledge Graph: park --founded in'--> year\\nThe Hot Ice Show --'hosted at'--> park
        Entity: 1. year 2. park
\end{tcolorbox}

\begin{tcolorbox}[title=Expand Entities, aclprompt]
Given the source entity "[source entity]" and the question below, generate new relations and objects start from source entity to expand the graph.

        Question: [question][sentence text]
        
        Step 1: List as much as relations what will help answer the question. Return relationships strictly in this format:
        
        [Tratectory]: your step-by-step thinking process.
        1. [relation1]
        2. [relation2]
        ...

        Step 2: Generate the object for each relation strictly in this format:
        
        [Tratectory]: your step-by-step thinking process.
        1. [source entity] -> [relation1] -> [object1]
        2. [source entity] -> [relation2] -> [object2]
        3. ...
        
        Requirements:
        
        1. All the relations and entities must be shown in natural laguage, without any special characters.
        2. **The object should be as specific as you can. For example, you should generate "Franois de Malherbe" not "Specific Secretary's Name", "Fogg" not "Specific Winner's Surname". You should avoid noisy and unmeaningful output.**

\end{tcolorbox}

\subsection{External Retrieval}

\begin{tcolorbox}[title=Select Triplet and Action, aclprompt]
You are given a question and a knowledge graph.

[QUESTION]: [question]

[KG]:
[Graph]

You can take one of two actions:

- Correct: Improve the correctness of a triplet that may contain errors.

- Expand: Add new related triplets based on existing ones.

You must select one **existing** triplet (i.e., an edge in the graph) that is **not marked [searched]**.

Return your decision in **exactly** the following format:

Action: [Enhance or Expand]

Triplet: Head --[Relation]--> Tail

**Strict Rules**:

- You must choose a triplet from the above [KG] block.

- Do NOT invent or merge triplets.

- Do NOT include more than one triplet.

- Use the **exact string match** from the KG.

- The format must be exactly: Head --[Relation]--> Tail

- Do NOT include explanation or justification.
\end{tcolorbox}

\begin{tcolorbox}[title=Correct Triplet, aclprompt]
You are given a knowledge triplet and a context paragraph. Your job is to revise this triplet if needed, based on the factual evidence in the context.

Triplet to correct:

[head] --[relation]--> [tail]

Context:

[retrieved context]

Only return ONE modified triplet in the following format:

Head --[Relation]--> Tail

- Use exact format with spaces and brackets as shown.

- Do NOT add any explanation, preamble, or extra text.

- Do NOT include "Answer:", "Output:", "Final answer:" etc.

- The returned triplet MUST appear in a single line exactly as described.

If no better version is found, you can return the original triplet as-is.
\end{tcolorbox}

\begin{tcolorbox}[title=Expand Triplet, aclprompt]
You are given a context paragraph related to this triplet:

[head] --[relation]--> [tail]

Your task is to extract UP TO 3 new factual triplets that are directly helpful to answer the following question:

"question"

Context:

[retrieved context]

Return your extracted triplets in the exact format:

Head --[Relation]--> Tail

- Only return factual knowledge expressed in the context.

- No explanation or commentary.

- Each triplet should be on a separate line, no bullet points or numbering.

- No extra text before or after.

Only include valid triplets. If no new triplets can be extracted, return an empty response.

\end{tcolorbox}

\subsection{Answer on Graph}
\begin{tcolorbox}[title=Answer on Graph, aclprompt]
You are a QA test machine, you need to answer the [Question] by a single entity.

A [Knowledge Graph] is given for your reference. Please find the single answer entity on the graph.

[QUESTION] [question]
[Knowledge Graph] [graph str]

Let’s think step by step to find answer entity in the knowledge graph. 

Requirements:

1. In your thinking trajectory, you should analyze what triplet does question look at, and how do you find proper answer entity on the given graph.

2. If there are multiple entities in the graph seem correct for question, you should choose the best one.

3. The answer entity is a single entity, it should be detailed and specific and you should not add any explanation after return the entity

4. If the question is a yes/no question, you should answer it by "yes" or "no"

\end{tcolorbox}

\subsection{Text-based RAG baseline}
\begin{tcolorbox}[title=Text-based RAG baseline, aclprompt]
You are a QA test machine, you need to answer the [Question] refer to the given [CONTEXT] and your knowledge. Let’s think step by step, and please
output the answer to the [Question] strictly follow the format of: 

Thinking trajectory: String

Final Answer: String.

[QUESTION] [question]

[CONTEXT] {retrieved docs}

\end{tcolorbox}

\end{document}